\documentclass[final,3p,times,twocolumn,authoryear]{elsarticle}

\usepackage{amssymb}
\usepackage{amsmath}
\usepackage{natbib}
\usepackage{graphicx}
\usepackage[export]{adjustbox}
\usepackage{lscape}
\usepackage{gensymb}

\journal{arXiv}

\begin{document}
	
	This work has been submitted to the IEEE for possible publication. Copyright may be transferred without notice, after which this version may no longer be accessible.

\begin{frontmatter}

\ead{huw.jones@manchester.ac.uk}

\title{Data-driven Approaches to Surrogate Machine Learning Model Development}

\author{H. Rhys Jones}
\author{Tingting Mu}
\author{Andrei C. Popescu}
\author{Yusuf Sulehman}

\address{University of Manchester, Oxford Rd, Manchester M13 9PL, United Kingdom}

\begin{abstract}

We demonstrate the adaption of three established methods to the field of surrogate machine learning model development. These methods are data augmentation, custom loss functions and transfer learning. Each of these methods have seen widespread use in the field of machine learning, however, here we apply them specifically to surrogate machine learning model development. The machine learning model that forms the basis behind this work was intended to surrogate a traditional engineering model used in the UK nuclear industry. Previous performance of this model has been hampered by poor performance due to limited training data. Here, we demonstrate that through a combination of additional techniques, model performance can be significantly improved. We show that each of the aforementioned techniques have utility in their own right and in combination with one another. However, we see them best applied as part of a transfer learning operation. Five pre-trained surrogate models produced prior to this research were further trained with an augmented dataset and with our custom loss function. Through the combination of all three techniques, we see an improvement of at least $38\%$ in performance across the five models.

\end{abstract}




\end{frontmatter}

\section{Introduction}


A machine learning surrogate (MLS) is a model which aims to explain natural or mathematical phenomena which can already be explained using an existing model. Using data from the original model, machine learning techniques are used to produce an optimised MLS model. The advantages of an MLS include increased computational efficiency when generating model outputs, with the trade-off being reduced accuracy. Once developed and trained, machine learning models (including an MLS) can produce new data instances almost instantly using a standard computer, whereas generating the same information using the original model and equivalent hardware may require hours or days of computational effort. The reduction in accuracy between an MLS and an original model must be quantified on a case-by-case basis and assessed on whether it is acceptable for practical use.

\begin{figure*}[h]
\centering
\includegraphics[scale=0.65]{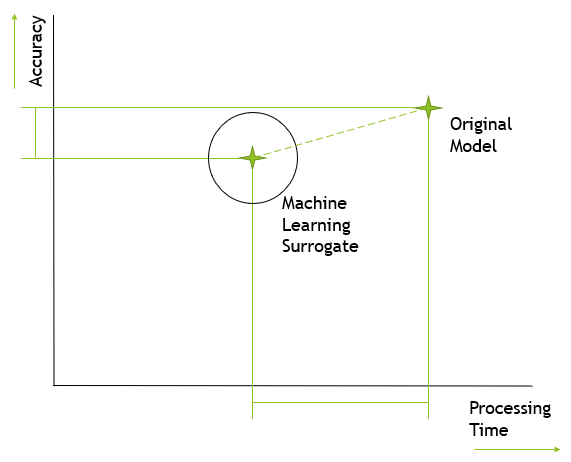}
\caption{The Trade-off Between a Machine Learning Surrogate Model and the Original Model. Once trained on data from the original model, the production of new data is likely to be significantly more efficient in terms of computation and time. However, as the machine learning model is produced using data from the original model, there will be some inevitable reduction in accuracy.}
\label{fig:surrogate_vs_model}
\end{figure*}

Previous research works have dealt with the production of MLS in areas such as material properties prediction [\cite{nyshadham2019machine}] and [\cite{asteris2021predicting}], with a recent work [\cite{jones2022surrogate}] focusing on seismic analysis for nuclear graphite cores. It is the MLS model from this latest research work that will be focused on in this paper. In the afforementioned works, a strong focus on neural networks [\cite{gurney1997introduction}] is seen, including convolutional neural networks (CNNs).

Despite the motivation for the production of MLS models being to reduce the need for expensive production of data, a large amount of this data is required to train such a model. A machine learning model trained on an insufficient number of data instances may result in overfitting [\cite{hawkins2004problem}]. Some techniques were employed in the aforementioned paper, including randomised layer dropout [\cite{srivastava2014dropout}], to counteract the effects of overfitting . 

A common technique used to improve model performance given a limited dataset is to manipulate existing data instances in a process known as data augmentation [\cite{perez2017effectiveness}]. This approach is commonly employed in machine learning applications involving image recognition and analysis [\cite{hansen2015tiny}], with techniques such as mirroring and rotation used to increase the number of data instances in a dataset.

Another commonly encountered problem during machine learning model development is dataset bias. In this situation, the dataset used to train the model is weighted towards a particular region of the input and/or output space. Alternatively, the dataset may be sparse in a particular region of the data space i.e. there may only be few data examples for a part of the data input or output continuum. Several methods can be employed to counteract the problem of dataset bias, including emphasising underrepresented data samples to a greater degree. We may instead use a loss function during model training which is designed to correct for dataset bias. 

A third problem encountered when training neural networks is the computational cost associated with their development and optimisation. This is particularly problematic when the problem space is complex - such as it is in this research. Instead of starting from scratch, we may use models produced from previous research works as a starting point during the development of neural networks for our own research. Through a process of transfer learning [\cite{torrey2010transfer}] we can adapt the model architecture, as well as the optimised weights, generated during previous works. By using transfer learning we may be able to make our model development process more efficient by reducing the time and computational resource needed to optimise a model for our purposes.   

A research question to be investigated and answered by this paper is whether data augmentation can be applied to problems such as machine learning surrogates. To this end, a framework will be developed to apply image manipulation techniques to the dataset used in the aforementioned graphite core model. In addition, we will investigate whether the use of custom loss functions and transfer learning can improve model performance.

\section{Background} \label{Background}

\subsection{Advanced Gas-cooled Reactors and the Parmec Model} \label{Parmec}

The computational model Parmec [\cite{wiki:xxx}] is the underlying model which was the subject of a MLS model in [\cite{jones2022surrogate}] and the same problem and base dataset is considered in this work. Parmec is employed to simulate the seismic response of the graphite core within the UK's advanced gas-cooled reactor (AGR). This model consists of a simplified 3-dimensional representation of the AGR graphite core, including the positional arrangement of the graphite bricks and other components. Parmec can be used to simulate a range of different seismic scenarios with the resulting component translation, rotation etc. being calculated by the model. 

In addition to the seismic configuration, another input to the Parmec model is configuration of cracked fuel bricks within the graphite core. Due to years of exposure to high temperatures and irradiation, some of the fuel bricks within the reactor are cracking, causing them to break into two pieces. The presence and configuration of these cracks has an impact on the reaction of the core to seismic loading. It is possible that up to 40\% of the fuel bricks will eventually crack, although it is difficult to determine or predict where and when cracks will occur.

The relationship between crack configuration and seismic response of core components is complex, hence the Parmec model consists of many thousands of parameters and equations. In addition, there are over $10^{2500}$ possible permutations of crack configuration, assuming 40\% cracking. With each configuration requiring around 2 hours to compute the seismic response via Parmec, it is clearly impractical to generate data for even a small percentage of them. Instead, industry practice is to generate random configurations of cracks, passing each through Parmec in order to build up a stochastic distribution of the seismic response.

\subsection{Previous Machine Learning Surrogate Model of Parmec} \label{previous}

\begin{figure}[ht]
\centering
\includegraphics[scale=0.4]{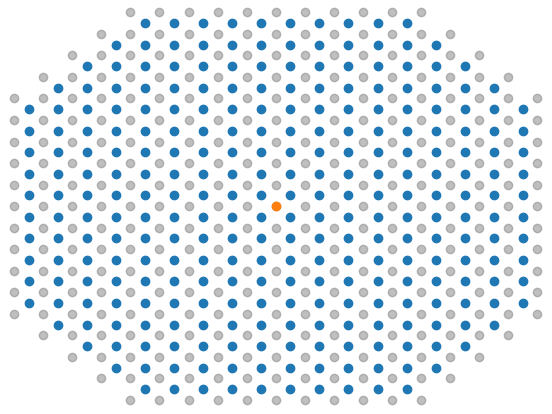}
\caption{A Top Down Diagram of the AGR Graphite Core Parmec Model. Bricks are arranged into channels of two different types: fuel (blue) and interstitial (grey). Both types of channel are the same height, with fuel bricks being stacked seven high and the shorter interstitial bricks being stacked 12 high. The cracking status of all 1988 fuel bricks is included in the input features (whether the brick is cracked or not) of the surrogate machine learning model. For the output labels, only the earthquake response of the upper most interstitial brick (orange) is predicted by the surrogate machine learning model. }
\label{fig:top_down}
\end{figure}

In previous machine learning assessments of AGR graphite core seismic analysis [\cite{jones2022surrogate}], each crack configuration is considered an individual data instance, with the encoding of cracked bricks being the input features and the response of core components to the earthquake being the output labels. The Parmec software generates a time-history of the earthquake response for all of the thousands of components within the core. For the sake of simplicity and focus, the MLS model was trained to predict the earthquake response for a single interstitial brick at a single time frame - see Figure~\ref{fig:top_down}. 

\begin{figure}[ht]
\centering
\includegraphics[scale=0.35]{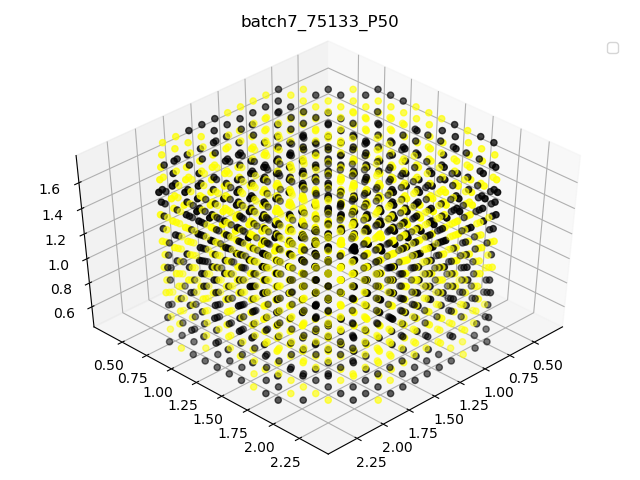}
\caption{Visualisation of a 3-dimensional Feature Encoding. This example represents a single instance with each data-point representing a fuel brick. Yellow and black data points represent uncracked and cracked bricks, respectively.}
\label{fig:cracked_3d}
\end{figure}

To summarise the features of the MLS, each instance has an input size of 1988 with this being the number of fuel bricks within the AGR graphite. This input was arranged into a 3D tensor which retains physical positional relationships within the actual AGR graphite core (Figure~\ref{fig:cracked_3d}). Each element is either a 1, -1 or 0 representing a cracked brick, uncracked brick or `empty' position. The 3-dimensional encoding of the input features also allows the dataset to be used with a convolutional neural network [\cite{albawi2017understanding}] which was found to be the best performing type of machine learning model.

For the aforementioned study, a dataset of approximately 8300 instances was created using the random crack pattern generator and the Parmec software. Out of these instances, 6300 (75\%) were used for training with the remaining 2000 samples retained for testing.                                       

\subsection{Data Augmentation} \label{augmentation}

Data augmentation is frequently employed in classification problems within the field of machine learning [\cite{shorten2019survey}], where the model predicts a discrete category for each dataset instance. A classic example of classification is in computer vision, where a 2D or 3D tensor representing an image is used to predict a category that is depicted. For example, models trained on the ImageNet dataset [\cite{deng2009imagenet}], which contains millions of images each representing one of 1000 discrete classifications, attempt to categorise the image depicted in an instance it is presented with. Classification is in contrast to regression problems, where there is a continuous, rather than discrete, output variable.

When dealing with problems such as ImageNet classification, performance is constrained by the size of the dataset. Model performance tends to improve with a larger number of training examples. A related constraint is dataset bias: regardless of the overall number of examples in the entire dataset, if one or more classes exhibits a significantly lesser or greater number of examples than the rest, model performance may be inhibited. Should there be a lack of examples for one particular class, not only will the model have difficulty identifying examples of this class, but performance for other classes will also be impacted.

Both of the aforementioned constraints tend to cause the phenomenon known as overfitting [\cite{hawkins2004problem}], where the model optimises too closely to the training data, including any noise or unrelated variability. There exist several methods to alleviate the effects of overfitting, including randomised nodal dropout [\cite{srivastava2014dropout}]. A commonly used solution to overfitting within image classification is data augmentation, where image manipulation techniques are used to generate additional data from existing examples. 

\begin{figure*}[h]
\centering
\includegraphics[scale=0.3]{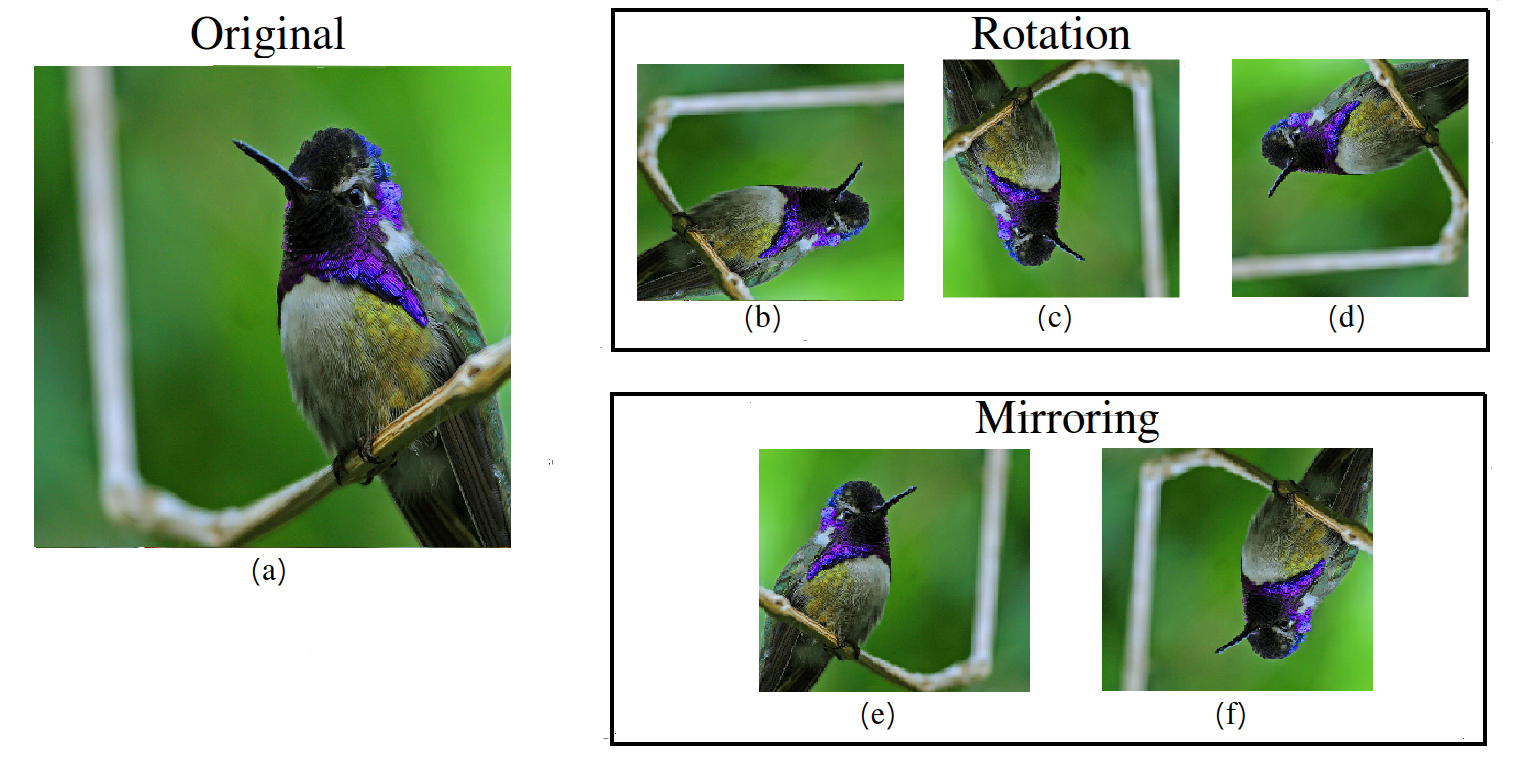}
\caption{An Example of Image Manipulation Techniques to Perform Data Augmentation. The base instance of an image depicting a bird is shown in image (a). The additional images show examples of two types of augmentation. Images (b), (c) and (d) show image (a) rotated by 90\degree, 180\degree and 270\degree, respectively. Similarly, images (e) and (f) show image (a) reflected about the vertical and horizontal axis, respectively. Despite being manipulated in this way, each image still effectively depicts an example of a bird and can be treated as such in the training of a machine learning model. Data augmentation can be used to expand a dataset without labelling additional examples, potentially improving model performance and reducing overfitting.}
\label{fig:data_augmentation}
\end{figure*}

 Figure~\ref{fig:data_augmentation} shows an example of a data augmentation process on a single image instance. The image on the left of this figure would correctly be classified as a bird. This image can be used to generate five additional instances of the same class: three by rotation and two by mirroring. By applying this process to all images within a dataset, the number of available training instances can be multiplied by a factor of seven.
 
 From the example, it can be seen that data augmentation techniques are highly suited to problems where the data is structured as a 2D or 3D tensor (such as greyscale or colour images, respectively). By extension, data augmentation is highly effective for applications in which localised or spatial patterns are of importance, for example where CNNs are employed. It should be noted at this point that the research work that we attempt to build on here employs 3D encoding of data as well as CNNs.

\subsection{Custom Loss Function} \label{LossFunction}

One of the most important model metrics to be selected during machine learning model development is the loss function. This function is used during training to calculate the difference between the ground truth and model prediction (known as the loss rate). Further, the derivative of the loss rate is used to iteratively update the model weights during the training and optimisation process.    

Typically, one of only a few loss functions will be selected for regression model training. A common selection is the mean squared error [\cite{wallach1989mean}] or one of its derivatives such as root mean squared error. Other options include mean absolute error [\cite{chai2014root}] and Huber loss [\cite{huber1964robust}] which was found to be optimal in the preceding research work on this topic [\cite{jones2022surrogate}].  

\begin{figure}[p]
\centering
\includegraphics[scale=0.5]{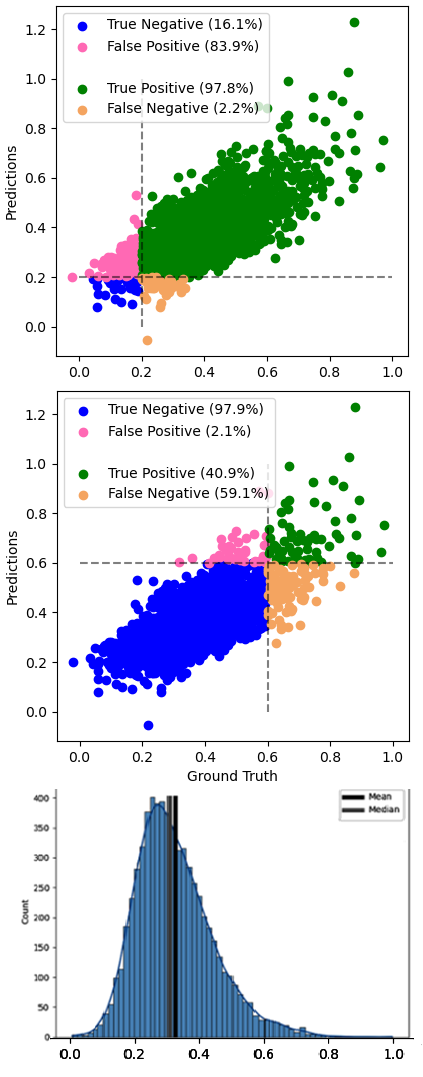}
\caption{The results from the best performing model produced via the method described in a previous research work [\cite{jones2022surrogate}]. This model is M3, the performance of which is listed in Table~\ref{tab:paper1results}. The top and middle images both show the model prediction plotted as a function of the ground truth values. For comparison, the distribution of the dataset is shown in the bottom image. Notice that beyond the delineations shown in the top and middle images (0.2 \& 0.6, respectively) there are fewer dataset examples, hence lower accuracy. }
\label{fig:data_distribution_results}
\end{figure}

An issue encountered in the aforementioned preceding research was that the data was not evenly distributed throughout the data space. The output data generated by the Parmec model tends to be distributed around a central modal value, with increasingly fewer examples towards the extremes of the data space. This in turn results in a model which tends to over-predict values in the lower part of the data space, and under-predict those in the upper part   (Figure~\ref{fig:data_distribution_results}).

This problem can be compared with the issue of class imbalance encountered in the field of machine learning classification [\cite{johnson2019survey}]. Much literature has been written on the subject of correcting for data imbalance in classification, with a recent work [\cite{sarullo2019class}] using a weighted loss function. 

Conversely, for regression based problems, research attention has been scarce by comparison. Some recent works [\cite{branco2017smogn}][\cite{yang2021delving}] note the lack of research on imbalanced regression have proposed solutions to problems caused by gaps or rarefactions in the data space. The solutions proposed involve the application of smoothing or dataset resampling. We note that the imbalance featured in these research works is not of the relatively smooth trend seen at the bottom of Figure~\ref{fig:data_distribution_results}.  

Whereas the data distribution in the aforementioned research papers contain discontinuities and irregular gaps, our dataset follows a regular pattern. This makes the methods explored previously in this area potentially unsuitable for the research problem at hand. If we wish to counteract the data imbalance problem in the research at hand, a bespoke custom loss function will have to be developed.

\subsection{Transfer Learning} \label{TransferLearning}

Transfer learning is a machine learning technique that focuses on using already trained models to serve a purpose beyond their original intent. Often in machine learning, models are trained from scratch, a process that consumes significant resources in terms of time and computation to achieve optimal performance. Transfer learning can serve to make the process more efficient and less resource intensive by using the knowledge from the pre-trained models in the training process.

In [\cite{jones2022surrogate}], an optimal model architecture was developed for the purpose of surrogation of a nuclear engineering model. Using this architecture and the training data, the parameters of the model were optimised from random starting weights. As training started from random weights, the process of model optimisation was repeated multiple times. The best performing models from this study, along with their pre-trained weights, can be transferred to this study as a starting point for exploitation of the methods described in subsections~\ref{augmentation}~\&~\ref{LossFunction}.   

\section{Preparation} \label{data}

\subsection{Image Manipulation Techniques} \label{manipulation}

\begin{figure*}[h]
\centering
\includegraphics[scale=0.5]{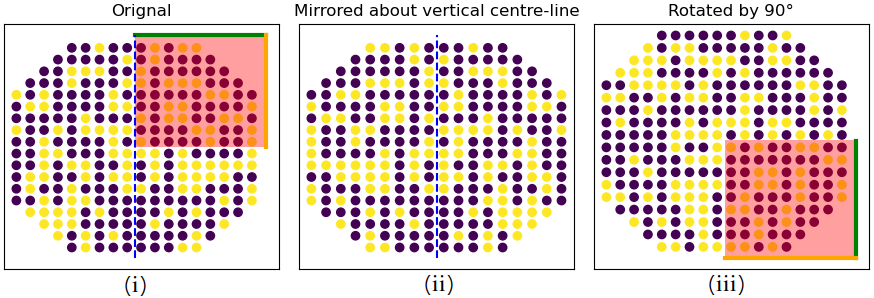}
\caption{An Example of Image Manipulation Techniques Applied to Parmec Data to Facilitate Data Augmentation. \textbf{i:} a slice from the feature inputs for an example instance: the dark blue spots represent intact fuel bricks, with the yellow spots being cracked bricks. This image is the equivalent of image (a) from Figure~\ref{fig:data_augmentation}, i.e. it is an original, unaltered instance. \textbf{ii:} the same example instance as shown in (i), but it has been mirrored about the vertical centre-line. This image is the equivalent of image (e) from Figure~\ref{fig:data_augmentation}. \textbf{iii:} again, the same example as (i), but this time it has been rotated by 90\degree - the equivalent of image (b) from Figure~\ref{fig:data_augmentation}. In each case, the output label effectively remains the same, as it represents the central brick of the core, about which the rotation or mirroring is performed.}
\label{fig:parmec_augmentation}
\end{figure*}

Recall from Figure~\ref{fig:cracked_3d} that the input features for this research problem is a 3D tensor representing the position of cracked fuel bricks within a given instance. This tensor is comparable to that of a colour image such as the one shown in Figure~\ref{fig:data_augmentation}. Being a tensor of a similar encoding to that in the aforementioned figure, the same image manipulation techniques can also be applied to the tensor for this research problem.

Recall also from Figure~\ref{fig:top_down} that the output labels for this research problem are continuous values representing displacement in the central brick of the core. Note also from the aforementioned figure that the overall data space from which the output variables are extracted is planar and can be expressed as a 2D tensor. For a given instance i.e. input/output  pair, we can apply any of the rotational or mirroring techniques shown in Figure~\ref{fig:data_augmentation}. Note that if we rotate or mirror about the vertical axis at the centre-point of the core, the encoding order of the input features will change, but the output label will not (as it represents the top brick in the central channel). Hence, data augmentation for the dataset in this study will create new instances with restructured feature tensors, but the same output label value - see Figure~\ref{fig:parmec_augmentation}. 

If we apply rotation (90\degree, 180\degree and 270\degree) and mirroring (vertical and horizontal) to all examples in our dataset, we can multiply the available number of instances by a factor of six. 

\begin{figure}[h]
\centering
\includegraphics[scale=0.8]{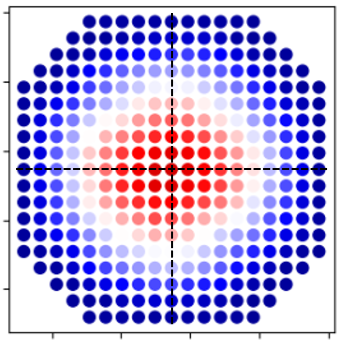}
\caption{Parmec Outputs for all bricks in the top level of the core. The colour of each brick data-point represents the average across the entire dataset (blue to red represents low to high, respectively). Note the symmetry about the vertical and horizontal centre-lines.  }
\label{fig:composite}
\end{figure}

As mentioned in subsection~\ref{augmentation}, the motivation behind using image manipulation techniques to create an augmented dataset is based on the conjecture that symmetries do exist in the Parmec data space. What is the justification behind the belief that rotational and symmetric manipulations of Parmec inputs would yield similarly transformed outputs? There are three lines of evidence that can be used to support the validity of this process:

\begin{enumerate}
    \item \textbf{AGR Design:} The design of the AGR (and the Parmec model that is based on it) contains four-fold symmetry [\cite{nonbol1996description}]. This effectively means that each quarter of the AGR (and Parmec) model is a rotation or mirror of the others. However, this is an incomplete justification as the simulated earthquake always impacts the model on one particular point on its periphery i.e. it is not symmetric.
    \item \textbf{Dataset Observation:} Observe Figure~\ref{fig:composite} which shows the average output value for each brick in the top layer of the Parmec model. It can be seen that there is a symmetry across both the vertical and horizontal centre-lines. This pattern is of the same four-fold symmetry as mentioned in the first bullet-point.
    \item \textbf{Augmented Equivalent Data:} Through the Parmec software package, we have the benefit of creating ground truth equivalents of augmented data. For a given ground truth example, we can manipulate the inputs according to one of the transformations shown in Figure~\ref{fig:parmec_augmentation} and then feed them through the Parmec model. Then we can compare the outputs generated by Parmec and our non-Parmec data augmentation technique.  

\end{enumerate}

We performed the process described in bullet point 3 above for two base instances, applying each of the five manipulation techniques on the inputs and then using these to generate labelled examples using Parmec. Simultaneously, we apply all five of our non-Parmec data augmentation techniques to the inputs and outputs of both examples. Comparing the outputs of both techniques, we notice agreement when applying rotation by 180\degree and when mirroring about the horizontal axis. 

The validity of the augmentation technique discussed here will ultimately be tested through an experimental machine learning process. We can train two machine learning models of the exact same architecture and parameters: one with a dataset augmented with and one with the base dataset. A separate testing set of only non-augmented instances will be retained for testing both models which both models can best tested against. The performance of each image manipulation method can be evaluated in this way.

\subsection{Weighted Loss Function} \label{Weighted}

\begin{figure}[h]
\centering
\includegraphics[scale=0.35]{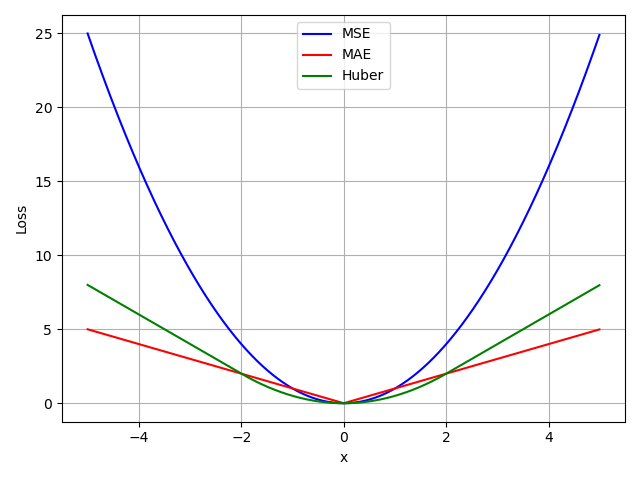}
\caption{Loss Functions: Visual Comparison. Three loss functions are compared graphically. \textbf{Blue}: mean squared error, as values become more extreme, the loss value increases geometrically meaning that outliers heavily influence the calculated value. \textbf{Red}: mean absolute error, the loss value increases linearly as the input increases, reducing the effect of outliers. \textbf{Green}: Huber loss, a balance between the previous two loss functions mentioned previously. The loss function has a linear outer region and a central non-linear region. }
\label{fig:lossPlots}
\end{figure}

In [\cite{jones2022surrogate}], the effectiveness of using three alternative loss functions were compared. These were the mean squared error (MSE), mean absolute error (MAE) and the Huber loss [\cite{huber1964robust}].
A model trained using the Huber loss function was found to produce the best performance. However, the other two loss functions produced a similar, albeit poorer, performance. Each of the three loss functions (Figure~\ref{fig:lossPlots}) each have their own strengths and weaknesses, with MSE heavily weighting outlying values, MAE proportionately weighting outliers and Huber being somewhere in between. As mentioned in subsection~\ref{LossFunction}, our base dataset is highly centred around a central value with a double tailed distribution. Regardless of the loss function used, the resulting model is biased towards the central region, resulting in difficulty predicting at the upper and lower extremes (Figure~\ref{fig:data_distribution_results}). 

We propose a loss function tailored to this dataset which applies an adjustment factor that is a function of model prediction distance from a central value.

\begin{equation}
Loss = \frac{\alpha^{2}}{n\beta^{2}} \sum_{i=1}^{n} Z_i^{2}(y_i - \Tilde{y}_i)^2.
\label{AdjustedLoss}
\end{equation}

\begin{equation}
Z_i =\frac{1}{\sigma \sqrt{2\pi}} \exp\big(-(\Tilde{y}_i - \mu)^{2}/2\sigma^2\big).
\label{Zterm}
\end{equation}

\begin{equation}
\beta \coloneqq \max\{Z_{i}: i=1, 2, \ldots, n\}.
\label{b}
\end{equation}

\begin{figure}[h]
\centering
\includegraphics[scale=0.45]{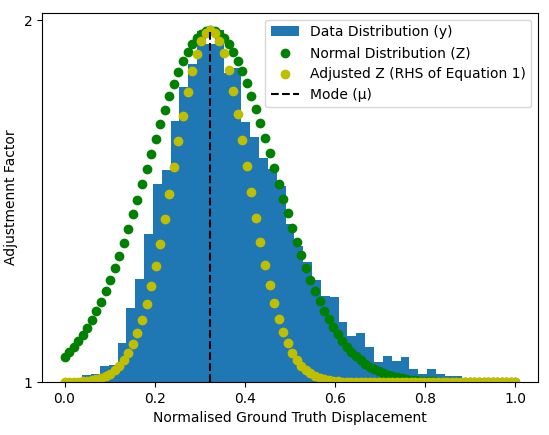}
\caption{Adjustment Factor as is Utilised by (\ref{AdjustedLoss}). The normal distribution (Z) is centred on the mode ($\mu$) and scaled by the standard deviation ($\sigma$) of the data distribution. This factor can be adjusted as per the right hand side of (\ref{AdjustedLoss}). }
\label{fig:adjustment}
\end{figure}

As can be seen from (\ref{AdjustedLoss}), the regular loss, which takes the mean of the square difference between the ground truth $y_i$ and the model prediction $\Tilde{y}_i$. This is then adjusted by the factor $Z_i$, as given by (\ref{Zterm}) and is calculated for each instance $i$. This term is based on the probability density function which in this case describes a Gaussian distribution. This Gaussian distribution is fit to that of our data distribution (Figure~\ref{fig:adjustment}) by use of the mode $\mu$ and standard deviation $\sigma$. This summation is scaled through division by the square of $\beta$, which denotes the maximum such $Z_i$. Also included is the magnitude coefficient $\alpha$ which will be optimised experimentally.

This function results in a strong adjustment for predictions made near the mode, quickly dropping off to one as we move away in either direction. This adjusted loss function penalises predictions made near the region where there is a large concentration of training data and instead pushes it towards the extremes. This equation is designed with the intention of counteracting the bias in the data distribution.

\subsection{Pre-Trained Model Transfer} \label{TransferModels}

In [\cite{jones2022surrogate}], multiple machine learning models were developed and trained on the dataset. In the best performing case, a convolutional neural network was utilised with a refined architecture. Optimal performance was achieved when including inputs representing the cracking status of the top 3 levels of the AGR core. The model was used to predict displacements in a single brick on the top level of the core at a single time point during an earthquake - the distribution of these outputs for the dataset can be seen in Figure~\ref{fig:adjustment}.  

Out of all of the models produced in the aforementioned paper, the best performing five models were obtained as well as their optimised weights. We evaluated each of these models, numbered M1 to M5, against a testing dataset with the results shown in Table~\ref{tab:paper1results}.

\begin{table}[h!]
    
    \begin{tabular}{c|c|c|r|c} 
      \textbf{Model} & \textbf{Test Performance (MSE) }  \\

      \hline
       & \\
      \textbf{M1}   &  9.28e-3 \\
      \textbf{M2}         & 9.25e-3 \\ 
      \textbf{M3}    & 9.22e-3  \\
      \textbf{M4} &    9.68e-3 \\
      \textbf{M5} &     9.48e-3  \\

      \end{tabular}
    \caption{Performance of the Five Best Performing Models Produced Using the Method of [\cite{jones2022surrogate}]. Each model was evaluated using a dedicated testing set sequestered for this research work. The mean squared error (MSE) metric was used to evaluate each model against this testing set.}
  \label{tab:paper1results}
\end{table}

These models will form the basis of transfer learning experiments through refinement using the methods described in subsections~\ref{manipulation}~\&~\ref{Weighted}. As mentioned in subsection~\ref{TransferLearning}, transfer learning can reduce the time and computational resources required compared to starting from randomised weights. Transferring models and weights in this case will not only reduce resource requirements but also be interesting from a research perspective. 

\section{Experimental Evaluation Process} \label{method}

\subsection{Augmentation} \label{augmentMethod}

We began by evaluating the effectiveness of each image manipulation technique mentioned in subsection~\ref{augmentation}. To do this, we augmented the base dataset using each of the five image manipulation techniques. This resulted in six available data sets: three using rotation, two using mirroring and the original unaugmented set. A summary of the datasets is shown in Table~\ref{tab:Augdatasets}.

\begin{table}[h!]
    
    \begin{tabular}{c|c|c|r|c} 
      \textbf{Dataset No.} & \textbf{Description }  \\

      \hline
       & \\
      \textbf{D0}   &  Original \\
      \textbf{D1}         & Rotation 90\degree \\ 
      \textbf{D2}    & Rotation 180\degree  \\
      \textbf{D3} &  Rotation 270\degree  \\
      \textbf{D4} &      Mirror Vertical  \\
      \textbf{D5} &      Mirror Horizontal
                     \\ 
            
      \end{tabular}
    \caption{Summary of the Datasets Used in Experiments 1 \& 2 as described in subsection~\ref{augmentMethod}. The base dataset (D0) is used to generate each subsequent dataset (D1 - D5) using the image manipulation techniques detailed in subsection~\ref{manipulation}. Each dataset is 6136 instances in size.  }
  \label{tab:Augdatasets}
\end{table}

For the purposes of comparison, a model design was selected which is highly simplified compared to that utilised in [\cite{jones2022surrogate}]. This simplification was made in order to reduce computational demands and to allow obtainment of results quickly. As our intention at this point is to compare the effectiveness of using different datasets and not overall optimisation, a simplified model is acceptable for this purpose. 

\begin{figure}[h]
\centering
\includegraphics[scale=0.35]{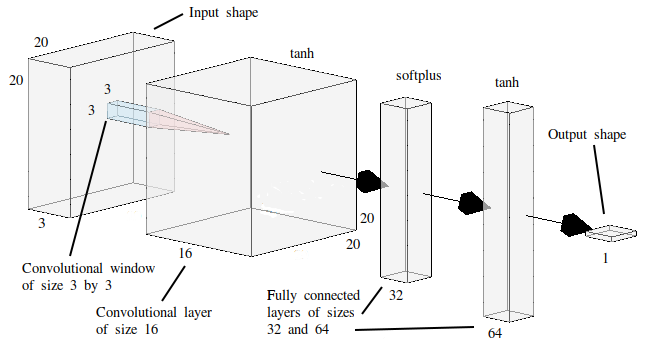}
\caption{ Simplified Model Architecture Used to Compare Augmentation Approaches. The input layer requires a 3-dimensional tensor representing the bricks of the top three layers of the AGR core. This is followed by a convolutional layer of 16 nodes, each with a 3x3 window size. Two fully connected layers then follow, the first with 32 nodes, the second with 64 nodes. These three layers have tanh, sofplus [\cite{zheng2015improving}] and then tanh again, respectively. Each of these layers utilised a 20\% dropout rate [\cite{srivastava2014dropout}] during training. The output layer represents only a single value - displacement in the single brick during the earthquake. This model is highly simplified compared to that of [\cite{jones2022surrogate}].  }
\label{fig:SimpleModel}
\end{figure}

The neural network architecture that was selected for this part of the research can be seen in Figure~\ref{fig:SimpleModel}. Between the input and output layers, there is one convolutional layer followed by two dense layers. Activation functions and the use of dropout was utilised based on previous experience. The Huber loss function was used for back propagation and optimisation during training.

\begin{table}[h!]
    
    \begin{tabular}{c|c|c|r|c} 
      \textbf{Experiment No.} & \textbf{Included Datasets }  \\

      \hline
       & \\
      \textbf{E1.0} &  D0 \\
      \textbf{E1.1} & D0 \& D1 \\ 
      \textbf{E1.2} & D0 \& D2 \\
      \textbf{E1.3} & D0 \& D3 \\
      \textbf{E1.4} & D0 \& D4 \\
      \textbf{E1.5} & D0 \& D5 \\ 
            
      \end{tabular}
    \caption{Summary of Experiment 1. Experiment 1.0 includes only the unaugmented dataset. Experiments 1.1 to 1.5 combine datasets D0 and one of the augmented datasets (D1 to D5). A description of each dataset is given in Table~\ref{tab:Augdatasets}. }
  \label{tab:Experiment1}
\end{table}

Six experiments were performed which involved training the model shown in Figure~\ref{fig:SimpleModel} individually with the datasets listed in Table~\ref{tab:Experiment1}. A 10\% sample of the unagumented dataset (D0) was retained for validation and the model was trained until reaching convergence in terms of validation loss. For each experiment, the training process was repeated 32 times, each time initialising with randomised starting weights and dropout nodes. Each model was then evaluated using a separate testing dataset with the results given in subsection~\ref{SingleAug}.

The datasets from the first phase were combined incrementally in the order of effectiveness as per Table~\ref{tab:Experiment2}. Again, each experiment is repeated 32 times. The results of this experiment are given in subsection~\ref{MultiAug}. 

\begin{table}[h!]
    
    \begin{tabular}{c|c|c|r|c} 
      \textbf{Experiment No.} & \textbf{Included Datasets }  \\

      \hline
       & \\

      \textbf{E2.1} & D0, D2 \& D4 \\           
      \textbf{E2.2} & D0, D2, D4  \& D1 \\      
      \textbf{E2.3} & D0, D2, D4, D1 \& D3  \\    
      \textbf{E2.4} & All \\
            
      \end{tabular}
    \caption{Summary of Experiment 2. The training set is expanded by combining datasets incrementally. The increments are performed in ranked order of effectiveness as per experiment 1.}
  \label{tab:Experiment2}
\end{table}

 \subsection{Custom Loss Function} \label{lossMethod}
 
 The purpose of this experiment is to evaluate the bespoke loss function defined in subsection~\ref{Weighted} and shown in (\ref{AdjustedLoss}) ~\&~(\ref{Zterm}). 
 The simplified architecture used in the previous section and shown in Figure~\ref{fig:SimpleModel} was again used in the experiments in this section. Initially, the unaugmented dataset (D0 from Table~\ref{tab:Augdatasets} was used for training and the same testing set as used in subsection~\ref{augmentMethod} for evaluation. 
 
 In addition, we will make adjustments to the (\ref{AdjustedLoss})~\&~(\ref{Zterm}) and train models using the same parameters. This will not only allow us to refine the function but also understand the impact of each component of it. In this part of the experiment, the alpha coefficient ($\alpha$) is set to unity.
 
 \begin{table}[h!]
    
    \begin{tabular}{c|c|c|r|c} 
      \textbf{Experiment No.} & \textbf{Loss Function }  \\

      \hline
       & \\

      \textbf{E3.1} & As per (\ref{AdjustedLoss})~\&~(\ref{Zterm}) \\ 
       & \\
      \textbf{E3.2} & Removing power of 2  \\  

       & from $\alpha, \beta$ and $Z_i$ of (\ref{AdjustedLoss}) \\
        & \\
      \textbf{E3.3} & Removing mean term ($\frac{1}{n}$) from   \\    
      & from (\ref{AdjustedLoss}) \\
       & \\
      \textbf{E3.4} & Combining E3.2 \& E3.3 \\
            
      \end{tabular}
    \caption{Summary of the Experiment 3. We begin by using the function detailed by (\ref{AdjustedLoss})~\&~(\ref{Zterm}) as the loss function (E3.1). We then remove the power of 2 term (E3.2). Next, we remove the mean term and instead take the full sum of the loss (E3.3). Finally, E3.4 combines both of the aforementioned term removals. In all four experiments, $\alpha$ is set to one.}
  \label{tab:Experiment3}
\end{table}

The next phase of research involved the use of our custom loss function in combination with the augmented datasets as detailed in subsection~\ref{augmentMethod}. All experiments outlined in Table~\ref{tab:Experiment4} use a training set which includes all augmented datasets as per E2.4. Further, we adjust the alpha coefficient in the remaining parts of this experiment, with $\alpha$ having a value of unity in E4.0 (as it was in experiment 3). E4.1, E4.2 \& E4.3 each increment $\alpha$ by unity in turn.

 As per experiments 1 \& 2, we will repeat each experiment 32 times to account for the stochastic nature of weight initialisation. A summary of the experiments performed in this section is given in Table~\ref{tab:Experiment3}. The results are reported in subsection~\ref{lossResults}.

 \begin{table}[h!]
    
    \begin{tabular}{c|c|c|r|c} 
      \textbf{Experiment No.} & \textbf{Loss Function }  \\

      \hline
       & \\

      \textbf{E4.0} & Dataset from E2.4 \& loss function  \\
      & from E3.1 \\
       & \\
      \textbf{E4.1} & As per E4.0 with $\alpha$ of 2  \\  

        & \\
      \textbf{E4.2} & As per E4.0 with $\alpha$ of 3 \\    

       & \\
      \textbf{E4.3} & As per E4.0 with $\alpha$ of 4 \\
            
      \end{tabular}
    \caption{Summary of the Experiment 4. We again use our custom loss function as defined by (\ref{AdjustedLoss}) \& (\ref{Zterm}) and evaluated in E3.1. However, this time we train on the full augmented training set as used in E2.4. In E4.0, we set $\alpha$ to unity, as it was in all parts of Experiment 3. In subsequent experiments (E4.1 to E4.3) we increase $\alpha$ by unity each time. }
  \label{tab:Experiment4}
\end{table}

 \subsection{Transfer Learning} \label{transferMethod}
 
 Subsection~\ref{TransferModels} discusses five pre-trained models obtained using the methodology of a previous research work in this field [\cite{jones2022surrogate}]. The performance of these models against a testing dataset is summarised in Table~\ref{tab:paper1results}.
 
The intention of this experiment is to further train these models using the methods detailed in the previous two subsections (\ref{augmentMethod} \& \ref{lossMethod}). We begin by further training models M1 to M5 using our dataset enlarged by all augmentation methods as per E2.4. We then combine both augmentation and the use of our custom loss function as defined in Equations\ref{AdjustedLoss} \& \ref{Zterm}. Finally, we perform the same further training of the transferred models using the conditions of the previous experiment but with an $\alpha$ coefficient of two.

This experiment is summarised in Table~\ref{tab:Experiment5}. We repeat the training process six times for each part of the experiment as opposed to the 32 times performed in earlier experiments. This is as only the dropout nodes are randomly selected and we are not initialising he weights. The results of this experiment will be presented in subsection~\ref{transferResults}.

\begin{table}[h!]
    
    \begin{tabular}{c|c|c|r|c} 
      \textbf{Experiment } & \textbf{Description }  \\
      \textbf{No.} & \\
      \hline
       & \\
       
      \textbf{E5.1} & Transfer learning with all  \\  
      
       &   augmentation datasets added\\

       &  to training set as per E2.4\\
       
        & \\

      \textbf{E5.2} & As per E5.1 but with the custom loss  \\  

        &  function defined by (\ref{AdjustedLoss})  \\
        
        & \\
      \textbf{E5.3} & As per E5.2 but with    \\    
      &  $\alpha$ coefficient of 2 \\

      \end{tabular}
    \caption{Summary of the Experiment 5. For each of our pre-trained models (M1 to M5 in Table~\ref{tab:paper1results}), we perform further training. We begin by enlarging the training set with all augmented datasets. We then combine this approach with the use of our custom loss function. Finally, we modify the loss function to use and $\alpha$ of 2. Each part of the experiment is repeated six times to account for randomness in the way dropout nodes are assigned.}
  \label{tab:Experiment5}
\end{table}

 \section{Results} \label{results}
 
 This section outlines several experiments used to test the hypotheses described in previous sections as well as a summary of the results.

\subsection{Augmentation} \label{augmentResults}

As mentioned in subsection~\ref{augmentMethod}, two augmentation experimental approaches are attempted. The first tests each image manipulation technique individually, the second exploits combinations of these techniques. The following two subsections report the results of these approaches, respectively.

\subsubsection{Individual Augmentation} \label{SingleAug}

The results of this experiment are summarised in Table~\ref{tab:Experiment1results}. The experiment which produced the optimal performance (i.e. lowest test loss) out of 32 repeats was E1.4 with 7.10E-3. This experiment also had the lowest mean loss (8.10E-3). All experiments which used an augmented dataset (E1.1 to E1.5) had a lower optimal performance than when using the unaugmented set only (E1.0). The mean values for all augmented experiments excluding E1.5 are below that of E1.0. 

\begin{table}[h!]
    
    \begin{tabular}{c|c|c|r|c} 
      \textbf{Experiment } & \textbf{Optimal Test} & \textbf{Mean Test}  \\
      
      \textbf{No.} & \textbf{Performance} & \textbf{Performance}  \\
      
      \hline
       & & \\
      \textbf{E1.0} & 1.06E-2 & 1.11E-2 \\
      \textbf{E1.1} & 7.70E-3 & 8.60E-3 \\ 
      \textbf{E1.2} & 7.20E-3 & 8.20E-3 \\
      \textbf{E1.3} & 7.90E-3 & 9.00E-3 \\
      \textbf{E1.4} & 7.10E-3 & 8.10E-3 \\
      \textbf{E1.5} & 1.02E-2 & 1.12E-2 \\ 
            
      \end{tabular}
    \caption{Results Summary of Experiment 1. Experiment 1.0 includes only the unaugmented dataset (D0). Experiments 1.1 to 1.5 combine datasets D0 and one of the augmented datasets (D1 to D5). The models produced during each experiment were tested against the testing set with the results reported in mean squared error (MSE). Out of the 32 models trained for each experiment, the optimal result (i.e. the lowest) is reported as well as the overall mean. }
  \label{tab:Experiment1results}
\end{table}

\begin{figure}[p]
\centering
\includegraphics[scale=0.275]{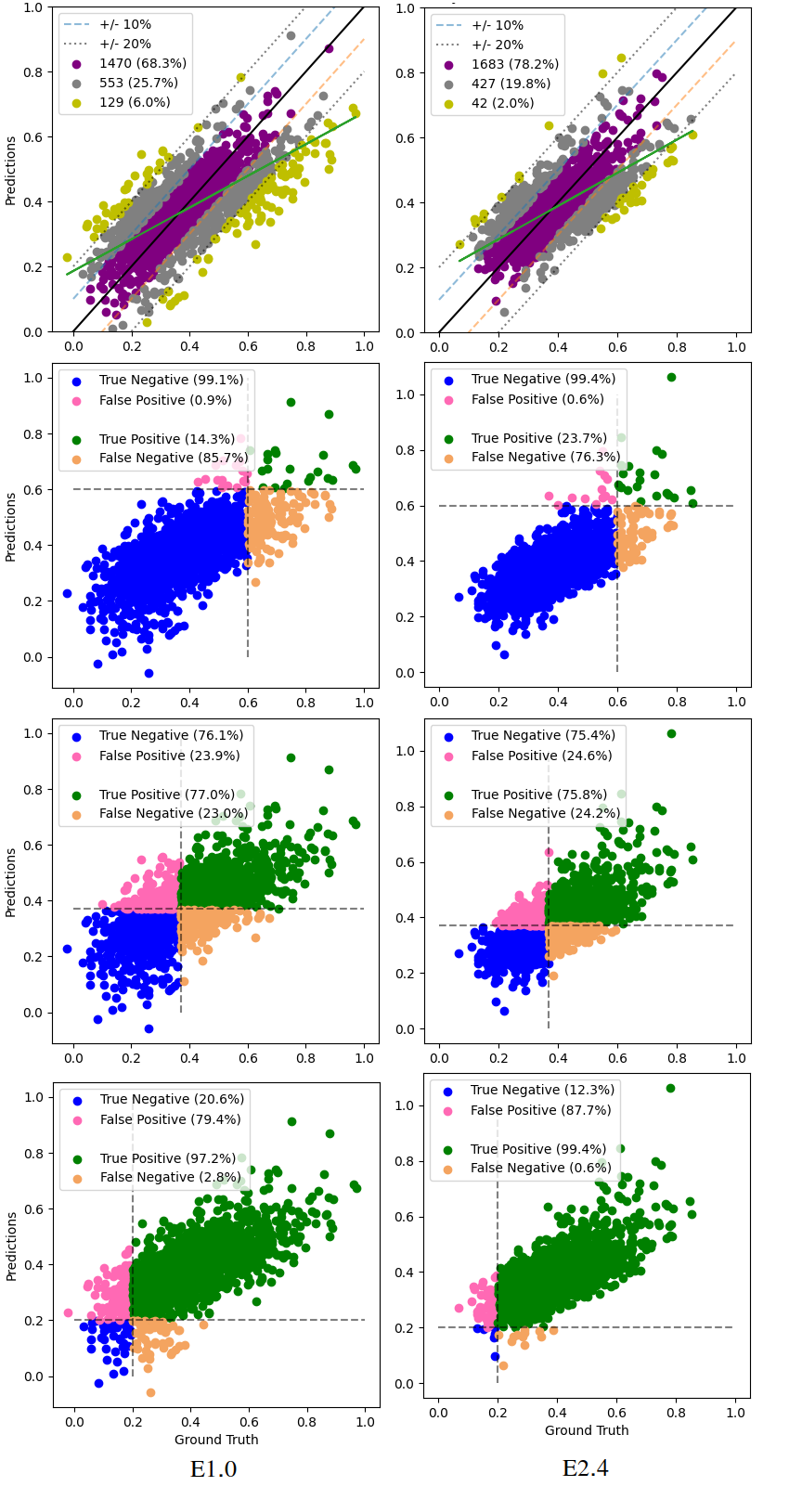}
\caption{Visual Summary and Comparison of the Performance of the Optimal Model Produced During E1.0 (\textbf{Left}) \& E2.4 (\textbf{Right}). The simplified model architecture shown in Figure~\ref{fig:SimpleModel} is trained on the original, unaugmented dataset. This process is repeated 32 times and evaluated against the test set. The predictions of the model with the lowest test loss are plotted against ground truth values and presented in four ways. In the top image, bounding lines are placed parallel to perfect prediction/ground truth agreement line (black), demarcating a 10 \& 20 percentage point margin. The lower three images split the data space into segments and quantify the proportion of samples which are correctly placed. }
\label{fig:E1p0_vs_E2p4}
\end{figure}

\subsubsection{Multiple Augmentation} \label{MultiAug}

The results of this experiment are summarised in Table~\ref{tab:Experiment2results}. All experiments which combine augmented datasets (E2.1 to E2.4) have improved optimal and mean test performances over that of the unagumented experiment (E1.0). As datasets are are incrementally added, both the optimal and mean test performances see improvement. 

In Figure~\ref{fig:E1p0_vs_E2p4} we see a visualisation and comparison of the test performance of the optimal model from E1.0 \& E2.4. 


\begin{table}[h!]
    
    \begin{tabular}{c|c|c|r|c} 
      \textbf{Experiment } & \textbf{Optimal Test} & \textbf{Mean Test}  \\
      
      \textbf{No.} & \textbf{Performance} & \textbf{Performance}  \\
      
      \hline
       & & \\
      \textbf{E1.0} & 1.06E-2 & 1.11E-2 \\
      \textbf{E2.1} & 7.30E-3 & 8.40E-3 \\  
      \textbf{E2.2} & 6.60E-3 & 7.70E-3 \\
      \textbf{E2.3} & 6.60E-3 & 7.70E-3 \\
      \textbf{E2.4} & 6.50E-3 & 7.50E-3 \\ 
            
      \end{tabular}
    \caption{Results Summary of Experiment 2. Each experiment combines the base unaugmented dataset (D0) with two or more augmented datasets as shown in Table~\ref{tab:Experiment2}. 
    Experiment E.10 is included for comparison and context. The models produced during each experiment were tested against the testing set with the results reported in mean squared error (MSE). Out of the 32 models trained for each experiment, the optimal result (i.e. the lowest) is reported as well as the overall mean.   }
  \label{tab:Experiment2results}
\end{table}

 \subsection{Custom Loss Function} \label{lossResults}
 
\begin{figure}[p]
\centering
\includegraphics[scale=0.45]{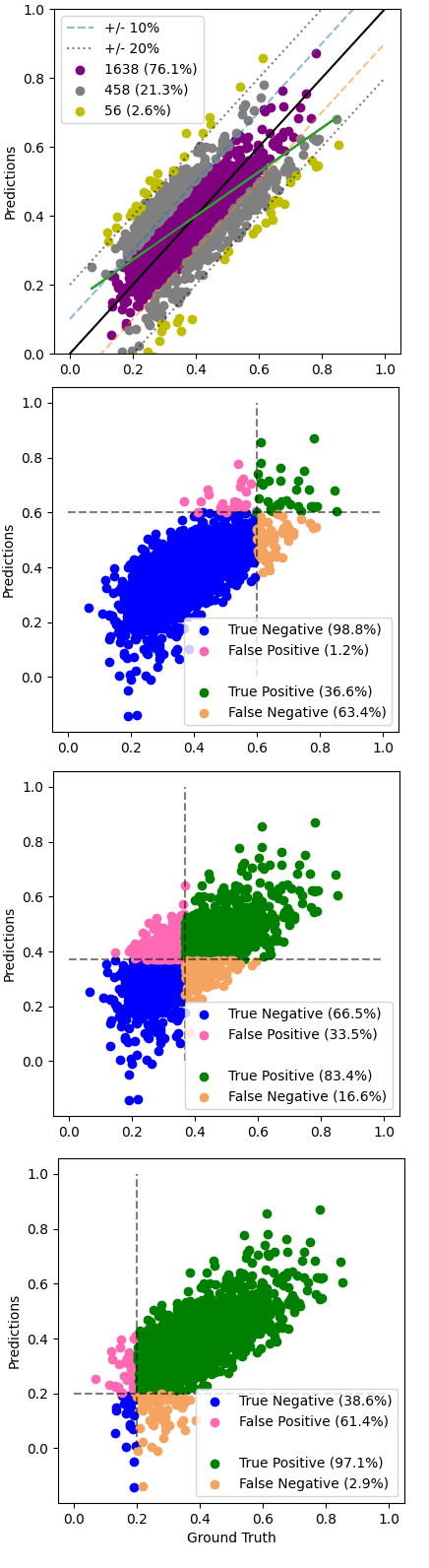}
\caption{Visual Summary of the Performance of the Optimal Model Produced During E4.1. The simplified model architecture shown in Figure~\ref{fig:SimpleModel} is trained on an expanded dataset which includes augmentation as described in subsection~\ref{manipulation}. The custom loss function as per (\ref{AdjustedLoss}) \& (\ref{Zterm}) is used during training with an $\alpha$ of value of 2. This process is repeated 32 times and evaluated against the test set. The predictions of the model with the lowest test loss are plotted against ground truth values and presented in four ways. In the top image, bounding lines are placed parallel to perfect prediction/ground truth agreement line (black), demarcating a 10 \& 20 percentage point margin. The lower three images split the data space into segments and quantify the proportion of samples which are correctly placed.}
\label{fig:e4p1_optimal}
\end{figure}

 An experimental approach to the development of a loss function customised to the needs of the data problem at hand was discussed in subsection~\ref{lossMethod} with a summary of proposed experiments shown in Tables~\ref{tab:Experiment3}~\&~\ref{tab:Experiment4}. The results are summarised in Tables~\ref{tab:Experiment3results} \& \ref{tab:Experiment4results} with a visual summary of E4.1 shown in Figure~\ref{fig:e4p1_optimal}. All parts of experiment 3 show similar mean squared error to that of the baseline case (E1.0) with little variation. The addition of all augmented datasets to the training set (E4.0) yields similar model performance to that of experiment E2.4 which uses the baseline loss function. Increasing the $\alpha$ coefficient above unity appears to increase mean squared error. 
 
 \begin{table}[h!]
    
    \begin{tabular}{c|c|c|r|c} 
      \textbf{Experiment } & \textbf{Optimal Test} & \textbf{Mean Test}  \\
      
      \textbf{No.} & \textbf{Performance} & \textbf{Performance}  \\
      
      \hline
       & & \\
      \textbf{E1.0} & 1.06E-2 & 1.11E-2 \\
      \textbf{E3.1} & 1.08E-2 & 1.11E-2 \\  
      \textbf{E3.2} & 1.06E-2 & 1.11E-2 \\
      \textbf{E3.3} & 1.08E-2 & 1.11E-2 \\
      \textbf{E3.4} & 1.07E-2 & 1.11E-2 \\ 
            
      \end{tabular}
    \caption{Results Summary of Experiment 3. Each experiment was trained on the base unaugmented dataset (D0) only. The results of E1.0, which involved training with a standard Huber loss function, are shown for context. Each part of the experiment involved a variation on the loss function shown in (\ref{AdjustedLoss}) \& (\ref{Zterm}). The models produced during each experiment were tested against the testing set with the results reported in mean squared error (MSE). Out of the 32 models trained for each experiment, the optimal result (i.e. the lowest) is reported as well as the overall mean.   }
  \label{tab:Experiment3results}
\end{table}

 \begin{table}[h!]
    
    \begin{tabular}{c|c|c|r|c} 
      \textbf{Experiment } & \textbf{Optimal Test} & \textbf{Mean Test}  \\
      
      \textbf{No.} & \textbf{Performance} & \textbf{Performance}  \\
      
      \hline
       & & \\
      \textbf{E4.0} & 6.54E-3 & 7.50E-3 \\
      \textbf{E4.1} & 7.31E-3 & 7.88E-3 \\
      \textbf{E4.2} & 8.14E-3 & 9.65E-3 \\
      \textbf{E4.3} & 4.17E-2 & 4.41E-2 \\

      \end{tabular}
    \caption{Results Summary of Experiment 4. Each part of this experiment involved training on the full augmented dataset as per E2.4 and the loss function from E3.1. We then increment the value of alpha. The models produced during each experiment were tested against the testing set with the results reported in mean squared error (MSE). Out of the 32 models trained for each experiment, the optimal result (i.e. the lowest) is reported as well as the overall mean.   }
  \label{tab:Experiment4results}
\end{table}

 \subsection{Transfer Learning} \label{transferResults}

We discussed in subsection~\ref{transferMethod} that our five existing models with their pre-trained weights (M1 to M5) were further trained using methods developed in this research work. The performance of these models further trained on our augmented training set is shown in Table~\ref{tab:Experiment5p1results}. Comparing these results with the original model performance (Table~\ref{tab:paper1results}) it can be seen that a significant improvement has been achieved.

Looking next at transfer models trained with the aforementioned augmented dataset and also using our custom loss function, very similar performance results are seen (Table~\ref{tab:Experiment5p2results}).

\begin{figure}[p]
\centering
\includegraphics[scale=0.275]{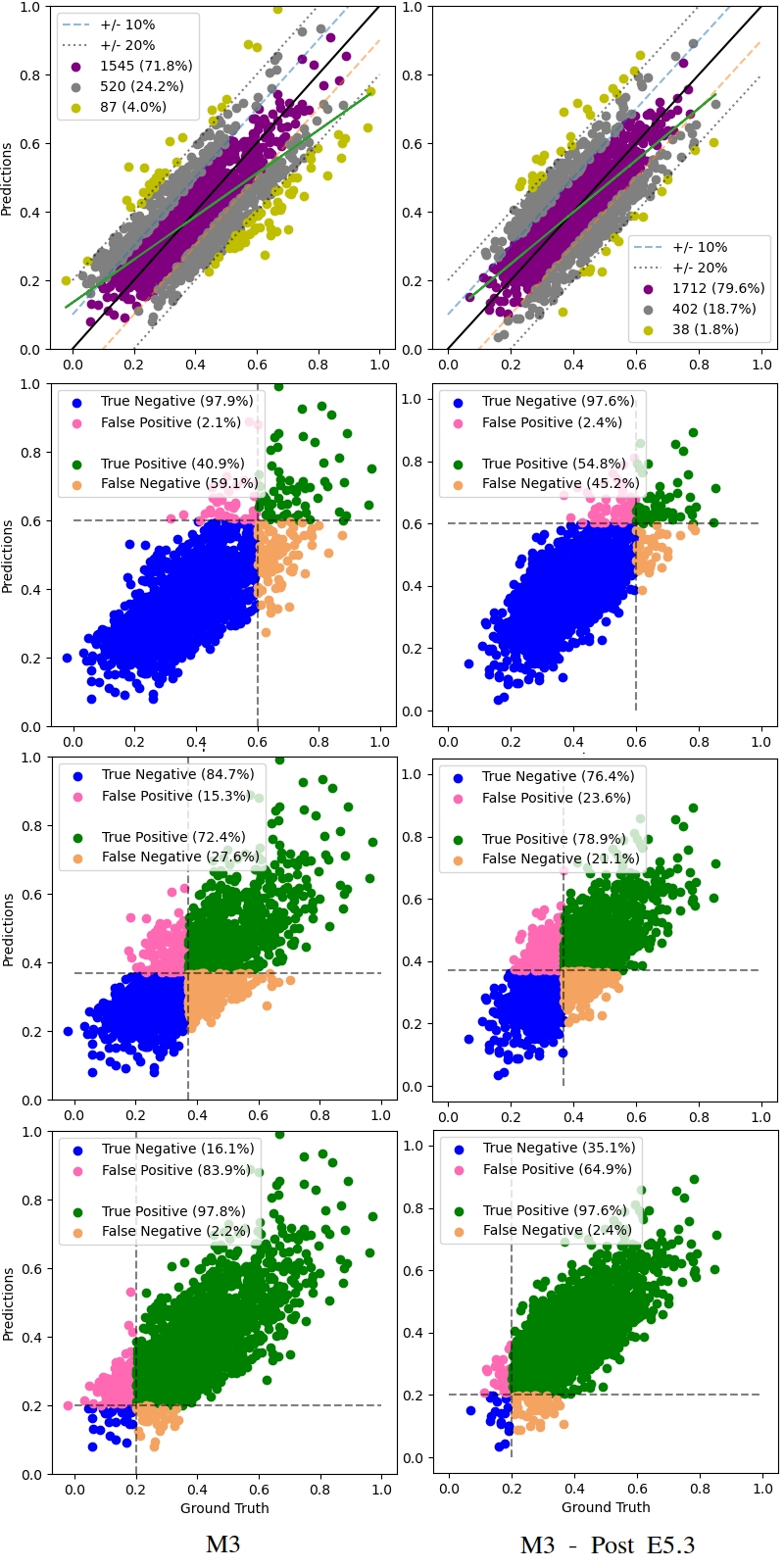}
\caption{Visual Summary and Comparison of the Performance of the Original Model M3 (\textbf{Left}) \& M3 After E5.3 (\textbf{Right}). The model obtained from [\cite{jones2022surrogate}] is trained using our custom loss function (\ref{AdjustedLoss}), with an $\alpha$ value of 2. This process is repeated six times and evaluated against the test set. The predictions of the model with the lowest test loss are plotted against ground truth values and presented in four ways. In the top image, bounding lines are placed parallel to perfect prediction/ground truth agreement line (black), demarcating a 10 \& 20 percentage point margin. The lower three images split the data space into segments and quantify the proportion of samples which are correctly placed. }
\label{fig:m3_vs_5p3}
\end{figure}

Finally, performing the same process as the aforementioned experiment but with the $\alpha$ set to two, we see the performance of the models summarised in Table~\ref{tab:Experiment5p3results} with a visual comparison with the original M3 model seen in Figure~\ref{fig:m3_vs_5p3} . Comparing E5.1 \& E5.2 with E5.3, it initially appears to perform slightly worse in terms of MSE test loss. From the comparison figure, multiple improvements over the original model performance can be seen, including a closer fit with fewer examples falling outside of both the 10 and 20 point margins. Also, the model from E5.3 performs considerably better than its original counterpart at the upper and lower bounds of the data space.
 
  \begin{table}[h!]
    
    \begin{tabular}{c|c|c|r|c} 
      \textbf{Model } & \textbf{Optimal Test} & \textbf{Mean Test}  \\
      
      \textbf{} & \textbf{Performance} & \textbf{Performance}  \\
      
      \hline
       & & \\
      \textbf{M1} & 5.70E-3 & 6.20E-3 \\
      \textbf{M2} & 5.70E-3 & 6.20E-3 \\
      \textbf{M3} & 5.70E-3 & 6.10E-3 \\
      \textbf{M4} & 5.70E-3 & 6.20E-3 \\
      \textbf{M5} & 5.80E-3 & 6.20E-3 \\

      \end{tabular}
    \caption{Results Summary of Experiment 5.1. This experiment involves the further training of the pre-trained models transferred from a previous study with an training set enlarged by data augmentation. The model performance detailed here should be compared with the original performance shown in Table~\ref{tab:paper1results}.   }
  \label{tab:Experiment5p1results}
\end{table}

  \begin{table}[h!]
    
    \begin{tabular}{c|c|c|c|c|c|} 
      \textbf{Model } & \textbf{Optimal Test} & \textbf{Previous Test} & \textbf{Reduction} \\
      
      \textbf{} & \textbf{Performance} & \textbf{Performance} & \textbf{\%} \\
      
      \hline
       & & & \\
      \textbf{M1} & 5.70E-3 & 9.28e-3 & 38.5 \\
      \textbf{M2} & 5.70E-3 & 9.25e-3 & 38.0\\
      \textbf{M3} & 5.70E-3 & 9.22e-3 & 38.0\\
      \textbf{M4} & 5.70E-3 & 9.68e-3 & 41.0\\
      \textbf{M5} & 5.80E-3 & 9.48e-3 & 39.0 \\

      \end{tabular}
    \caption{Results Summary of Experiment 5.2. This experiment involves the further training of the pre-trained models transferred from a previous study with an training set enlarged by data augmentation and also the use of a custom loss function. The model performance from this experiment is compared with that obtained previously in [\cite{jones2022surrogate}].  }
  \label{tab:Experiment5p2results}
\end{table}

 \begin{table}[h!]
    
    \begin{tabular}{c|c|c|c|c} 
      \textbf{Model } & \textbf{Optimal Test} & \textbf{Previous Test} & \textbf{Reduction}  \\
      
      \textbf{} & \textbf{Performance} & \textbf{Performance} & \textbf{\%} \\
      
      \hline
       & & & \\
      \textbf{M1} & 6.00E-3 & 9.28e-3 & 35.0 \\
      \textbf{M2} & 6.20E-3 & 9.25e-3 & 33.0\\
      \textbf{M3} & 6.40E-3 & 9.22e-3 & 30.5\\
      \textbf{M4} & 6.40E-3 & 9.68e-3 & 34.0 \\
      \textbf{M5} & 6.40E-3 & 9.48e-3 & 32.5\\

      \end{tabular}
    \caption{Results Summary of Experiment 5.3. Like E5.2, this experiment involves the further training of the pre-trained models transferred from a previous study with an training set enlarged by data augmentation. However, this time we the use a custom loss function with an $\alpha$ of two. The model performance from this experiment is compared with that obtained previously in [\cite{jones2022surrogate}].    }
  \label{tab:Experiment5p3results}
\end{table}

\subsection{Analysis and Discussion} \label{analysis}

Looking first at the augmentation experiments (see subsection~\ref{augmentResults}), we can see that all dataset augmentation has a positive effect on model performance. The results of experiment 1 (Table~\ref{tab:Experiment1results}) show that the addition of some augmented datasets have more of an effect than others. For example, experiments E1.4 \& E1.2 (rotation by 180\degree and vertical mirroring) both see about a one third reduction in mean squared error in comparison to the baseline (E1.0). It is interesting that the two best performing augmentations both make adjustments about the vertical axis. Conversely, the least performing augmented dataset (D5, included in E1.5) makes its adjustment about the horizontal axis.

It is clear from experiment 2 (Table~\ref{tab:Experiment2results}) that the combination of all augmented datasets to the training set yields the best result (E2.4). A deeper analysis of the performance of the model produced in experiment E2.4 can be made by comparing the left and right parts of Figure~\ref{fig:E1p0_vs_E2p4}. It can be seen that a model trained on all augmented datasets produces an overall better fit than when using no augmentation at all. Key indicators include the percentage of data within 20 points of the ground truth - 98\% for the best performing model from E2.4 compared with 94\% for a similar model from E1.0. The inclusion of all augmented datasets also appears to improve model prediction in the upper part of the data space - compare the second section from the top in Figure~\ref{fig:E1p0_vs_E2p4}. However, the best model from E2.4 also performs worse on predictions in the lower part of the data space (compare bottom parts of the aforementioned figure).

Recall that the model architecture used in these experiments is highly simplified compared to previous works in this field. We should note at this point that the results of experiment 2 are an improvement even over those of highly complex models from [\cite{jones2022surrogate}] - compare with the performance of M1 to M5, for example (Table~\ref{tab:paper1results}). This suggests that dataset size is far more important in this problem space than model complexity and refinement.

From an initial assessment of experiment 3 (Table~\ref{tab:Experiment3results}), it appears that the use a custom loss function as defined in Equations~\ref{AdjustedLoss} \& \ref{Zterm} has no advantage compared to the baseline (E1.0). All of the reported mean squared error test values are within a few percent of one another. 

Experiment 4 combined our custom loss function with all augmented dataset from experiment 2. A model trained with the combined custom loss function and full augmented data set (E4.0) has a similar performance to that of a model when using the augmented set alone (E2.4). Increasing the value of the coefficient $\alpha$ produces an increasing optimal and mean test performance.

So is there any value in the method evaluated in experiments 3 \& 4? To answer this question, we must return to the motivation behind this method as detailed in subsections~\ref{LossFunction}. We discuss the fact that our data space is concentrated in a central region with increasing rarefaction as we move away from it. Consequently, our models trained on this data performs poorly near the extremes as can be seen from the left part of Figure~\ref{fig:E1p0_vs_E2p4}. Comparing the aforementioned figure with a visualisation of the optimal model from experiment E4.1 ($\alpha$ = 2), we can see significant improvements in model performance at the upper and lower extremes of the data space (Figure~\ref{fig:e4p1_optimal}). This advantage comes at the expense of minor reductions in overall performance metrics, for example percentage of examples that are predicted outside of 20 percentage points of the ground truth (rising from 2.0\% in E2.4 to 2.6\% in E4.1). Whether or not this trade-off is of value will dependent on any practical application of the model. Nonetheless, it is likely that analysis using any such model would value good performance near the extremes as events in these regions are are most likely to impact safety. Therefore, we highlight the potential of this method and carry it forward into experiment 5.

We can see from the results of experiment 5.1 (Table~\ref{tab:Experiment5p1results}) that further training our pre-trained models with an augmented dataset results in a significant performance uplift. For example, M1 sees its test mean squared error drop from 9.28e-3 to a minimum of 5.70e-3 - a performance improvement of about 38.5\%. Combining the augmented dataset with our custom loss function and further training our base models appears to have little effect beyond what we saw in E5.1 (Table~\ref{tab:Experiment5p2results}). Repeating E5.3 with an alpha coefficient of 2 initially appears to worsen performance, with test MSE rising from 5.70E-3 to between 6E-3 and 6.40E-3. However, on inspection of the visual results of E5.3 and comparison with those of E5.1 \& E5.2, we see significantly improved performance at the upper and lower boundaries (above 0.6 and below 0.2 on the normalised scale of ground truth). Hence, we present the optimal results from E5.3 in Figure~\ref{fig:m3_vs_5p3}. Some notable improvements include a halving in the number of predictions that fall outside of the 20 percentage point margin (4.0\% in the original M3 against 1.8\% post E5.3), an increasing percentage of samples correctly placed over the 0.6 boundary (40.9\% in the original M3 against 54.8\% post E5.3) and finally a similar improvement below the 0.2 boundary (16.1\% against 35.1\%). As touched on earlier, in the field of machine learning for nuclear energy, this represents noteworthy progress as safety decisions are likely to concern events at the extremes i.e. very high or very low values.

\section{Conclusion} \label{conclusion}

In this research work, we demonstrate the adaption of three established approaches to the field of surrogate machine learning model development. The methods are data augmentation, custom loss functions and transfer learning. Each of these approaches have seen widespread use in the field of machine learning, however, here we apply them specifically to surrogate machine learning model development.

The machine learning model that forms the basis behind this work was intended to surrogate a traditional engineering model used in the UK nuclear industry. This model was built with the intention of increasing computational efficiency over the original model it surrogated. The performance of this model was hampered by poor performance due to limited training data. Here, we demonstrate that through a combination of additional techniques, model performance can be significantly improved.

Through exploitation of symmetry in the data and use of image manipulation techniques, we find that data augmentation techniques that make adjustments about the vertical axis are most effective, when applied individually. The combined use of all proposed augmentation methods in tandem produces the best performance uplift, suggesting that each of the methods provides at least some utility.

The second approach details an experimental refinement of a custom loss function specifically tailored to the training data distribution. We show that our custom loss function can improve performance at the extreme upper and lower parts of the data distribution - areas where previous models had performance difficulties.

We show that each of the aforementioned techniques have utility in their own right and in combination with one another. However, we see them best applied as part of a transfer learning operation. Five pre-trained surrogate models produced prior to this research were further trained with the augmented dataset and with our custom loss function. Through the combination of all three techniques, we see an improvement of at least $38\%$ in performance across the five models.

\section{Future Work}

Further work will aim to build on the methods developed in this research work. This includes refinement and fine tuning of the custom loss function developed in this paper. In addition, we will look at other techniques that could be used to improve model performance. One area with potential application to our research problem is active learning [\cite{ren2021survey}]. This technique uses additional feedback within the learning process and is often employed in situations where training data is in short supply. As we have established in this research work, techniques which compensate for a lack of training data or gaps in training data can be used to improve performance. Therefore, active learning is a worthwhile area to study next.

\section{Acknowledgments} \label{acknowledgements}

 The authors would like to acknowledge the assistance given by Research IT and the use of the Computational Shared Facility at The University of Manchester.

 We would also like to thank the teams behind the open-source projects NumPy [\cite{harris2020array}] and Matplotlib [\cite{Hunter:2007}], used widely throughout this project.

\bibliographystyle{elsarticle-harv} 
\bibliography{biblio}


\end{document}